# Accuracy and repeatability of a parallel robot for personalised minimally invasive surgery

Doina Pisla[1,5][0000-0001-7014-9431], Paul Tucan[1][0000-0001-5660-8259], Damien Chablat[1,4][0000-0001-7847-6162], Nadim Al Hajjar[2][0000-0001-5986-1233], Andra Ciocan[2][0000-0003-0126-6428], Alexandru Pusca[1][0000-0002-5804-575X], Adrian Pisla[1][0000-0002-5531-6913], Corina Radu[3][0000-0003-0005-0262], Grigore Pop[1][0000-0002-0557-368X] and Bogdan Gherman[1*][0000-0002-4427-6231]

[1]CESTER, Technical University of Cluj-Napoca, 400114 Cluj-Napoca, Romania
[2] Department of Surgery,"Iuliu Hatieganu" University of Medicine and Pharmacy, 400347 Cluj-Napoca, Romania
[3] Department of Internal Medicine, "Iuliu Hatieganu" University of Medicine and Pharmacy, 400347 Cluj-Napoca, Romania
[4] École Centrale Nantes, Nantes Université, CNRS, LS2N, UMR 6004, F-44000 Nantes, France
[5] Technical Sciences Academy of Romania, 26 Dacia Blvd, 030167 - Bucharest, Romania
*Corresponding author: `Bogdan.Gherman@mep.utcluj.ro`

**Abstract.** The paper presents the methodology used for accuracy and repeatability measurements of the experimental model of a parallel robot developed for surgical applications. The experimental setup uses a motion tracking system (for accuracy) and a high precision measuring arm for position (for repeatability). The accuracy was obtained by comparing the trajectory data from the experimental measurement with a baseline trajectory defined with the kinematic models of the parallel robotic system. The repeatability was experimentally determined by moving (repeatedly) the robot platform in predefined points.

## 1      Introduction

The advancements of classical surgery have seen notable progress due to the development of multiport or single -port surgical techniques.  These techniques represent a transition from traditional surgery, where the patient undergoes an incision ranging between 150 and 250 millimeters, to performing a reduced number of incisions (3 or 4) with a range varying between 10 and 25 millimeters in the case of multiport surgery, or a single incision varying between 10 and 25 millimeters in the case of single-port surgery (SILS) [1, 2]. These two techniques offer several advantages compared to classical surgery: the patient's recovery time is reduced, postoperative ileus experienced by the patient are significantly diminished, better cosmetics and there's a considerable reduction in blood loss [3, 4]. Alongside these advantages, these two techniques come with a series of limitations such as: collisions between instruments, the working space is reduced, lack of tactile feedback, low precision, visualizing the operative field using an endoscopic camera, diminished ergonomics and surgical hand dexterity, instrument crossing (in SILS) and, the surgeon's inability to manip-ulate more than two instruments simultaneously, and the position of the surgeon above the patient, culminating with prolonged intraoperative time [5, 6]. To reduce or eliminate these disadvantages, a series of robotic mechanical structures have been developed since 2000 when the first robot with FDA (Food and Drug Administration) approval is introduced in the field of surgery (da Vinci - developed by Intuitive Surgery). Intuitive Surgery holding a market monopoly with the structure such as: da Vinci S, da Vinci Si, and da Vinci SP until 2017 when a significant competitor emerged with the commercialization of a new multi-arm robot named Senhance, developed by Asensus Surgical [7, 8]. These structures are come with a series of advantages such as: improved medical ergonomics due to the use of the master-slave concept [9] (the surgeon controls the robot from a seated position through a master console), the surgeon's hand tremor elimination, reduced collisions between instruments, ability to control multiple instruments, improved accuracy and safety in medical procedures. The disadvantages of these system are: high cost of medical procedures, reduced intraoperative workspace, arm collisions, occupying a large volume in the operating room, the need to design new operating rooms adapted to the system's specifications, and the lack of tactile feedback need for out of sight instrument recognition [10].

To mitigate some of the previously mentioned disadvantages a new parallel robot for robotic assisted-surgery (PARA-SILSROB) [11] was developed. To validate the feasibility of the PARA-SILSROB robot experimental model, the robotic system accuracy and repeatability must be measured to determine if these parameters are within the values required for the medical tasks (to ensure safety and adequate control).

The paper presents the experimental assessment of PARA-SILSROB robot accuracy and repeatability using two methods: optical motion tracking and coordinate measuring. Starting from these methods, the accuracy, and the repeatability of the PARA-SILSROB robot is determined using the OptiTrack system [13] for optical detection and Stinger II [14] robotic arm for coordinate measuring. The robot's accuracy and repeatability are determined by comparing the trajectory of the robot's end-effector obtained using optical tracking and the coordinate measuring system with the trajectory generated in MATLAB using the kinematic model of the robot, thus functionally validating the mathematical model of the robot.

Following the Introduction section, the paper is structured as follows: Section II, Materials, and methods, presents the experimental setup of the PARA-SILSROB robot, the preliminary setup of the tracking equipment and methodology used to determinate the accuracy and repeatability of the system. Section III presents the results of experimental assessment and Section IV presents the conclusions and future work.

## 2 Materials and methods

There are different techniques and systems of tracking, acquisition and processing biomechanical data, often used in clinical research, as the systems based on wearable sensors [15, 16] or systems based on video cameras. To determine the accuracy and repeatability of the PARA-SILSROB, optical motion tracking and Stinger II measuring arm were used.

### 2.1 The experimental setup

The experimental model of the PARA-SILSROB robot was designed with respect to the medical protocol and surgeon's requirements [2, 12, 17], to control the laparoscopic camera and two active instruments used in minimally invasive surgery. The detailed presentation of the PARA-SILSROB robot's functionality, the experimental model, and the mechanical design of the robot are described in [2], [11], and [18] where the authors describe the robotic structure compose of three identical kinematic chains and the mobile platform developed to manipulate the active instruments and the laparoscopic camera necessary to perform the surgery. The experimental model of PARA-SILSROB robot is shown in Fig 1. Master-Slave concept [9] is used to control the robot, the main console can be equipped with two haptic devices or two 3D Space Mouses.

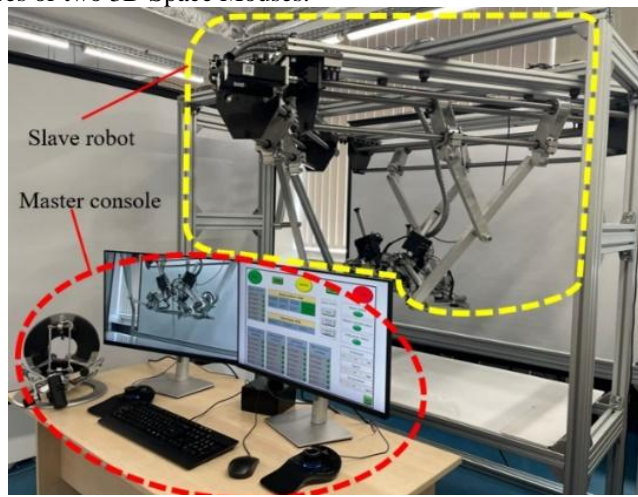

**Fig. 1.** The experimental model of the PARA-SILSROB.

The OptiTrack [13, 19] motion tracking system contains 6 cameras connected to a common switch controller that collaborates with a windows application installed on the computer. The cameras are placed around the robotic structure in positions that could achieve images of the optical markers placed on the mobile platform of the robot (Fig.2). This calibration of the system is performed with the help of the CWM 125 calibration device, guided manually in the OptiTrack visual field, each one recording the number of points based on which the workspace of the measurement system is generated. After the system calibration, a fixed reference frame (OXYZ) is defined using the CS-400 calibration square (Fig. 2).

Before placing the optical markers on the mobile platform of the PARA-SILSROB, the robot is initialized and placed in "Home" position (where the orientation angle on each axis is zero). Four markers were used for the measurements, placed on a smooth surface attached to the mobile platform; each marker has a diameter of 6.4 mm. The placement of the markers is shown in Fig 3.

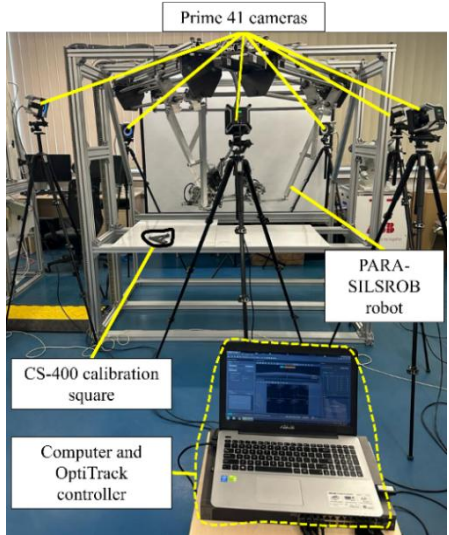 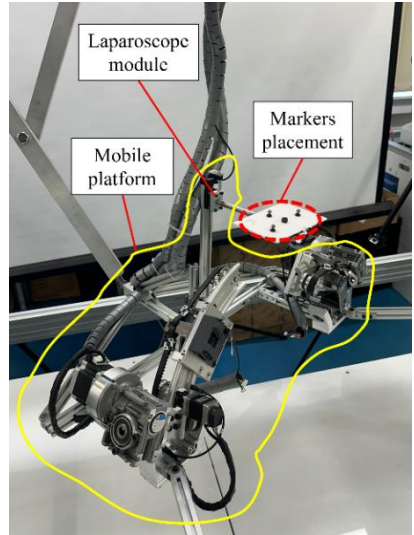

**Fig. 2.** Experimental setup for optical motion tracking.

**Fig. 3.** Optical markers position during the experimental trials.

Fig. 4 shows the position of the reference frame and the markers placed on the mobile platform of the robot within the application window of OptiTrack software.

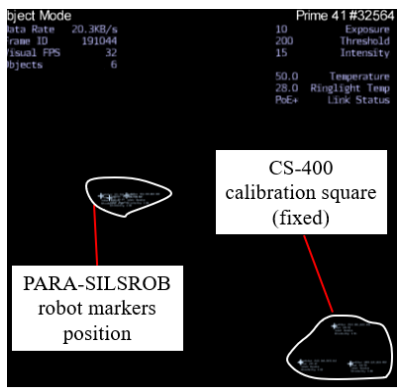 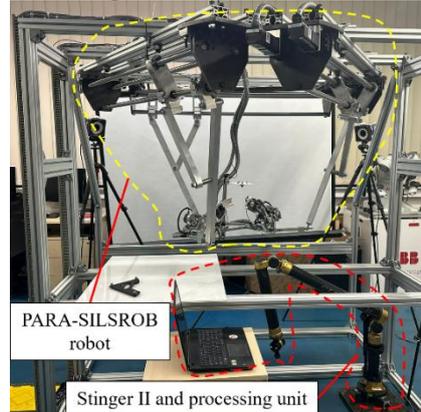

**Fig. 4.** PARA-SILSROB robot markers placement via Motive software.

**Fig. 5.** Experimental setup for coordinate measuring.

The experimental setup used to determinate the repeatability for coordinate measuring is presented in Fig. 5. The repeatability of the robotic system is determined by using the Stinger II measuring arm [14]. The advantage of this setup used for measurement is the fact that the Stinger II measuring arm does not require calibration, it becomes active after reading the positions on the axes.

### 2.2 Methodology

Fig.6 illustrates the methodology defined for determining the accuracy, and Fig. 7 illustrates the methodology of determining the repeatability of the parallel robotic system. After performing the experimental measurements using the optical tracking system, results were extracted from the systems database and imported in MATLAB for further processing. The OptiTrack system provides an exportable CSV file containing time-based values of the optical markers coordinates (*X*, *Y* and *Z*). In order to allow an accurate reading of the coordinates, a 2 seconds delay was imputed between the motion command and the motion start of PARA-SILSROB, to assure that the starting time of the optical tracking system is the same with the starting time of the motion. The robot performs a motion using three points, from the starting point, the mobile platform is moved to point 1 then to point 2 from where it moved back to starting position. Three sets of coordinates are obtained for each marker placed on the mobile platform (*X*, *Y* and *Z*). The markers are placed on the mobile platform so that they form a tringle with the centroid (computed using Eq. 1) coinciding with the tool center point (TCP) of the mobile platform. Using MATLAB, the same motion characteristics used in the experimental setup for the robot motion are imputed in a trajectory computation program based on the kinematic modelling of the robot in order to obtain the virtual trajectory of the TCP. The results obtained after processing the data from OptiTrack are compared with the data from the trajectory computation algorithm from MATLAB and graphically represented using the same reference system.

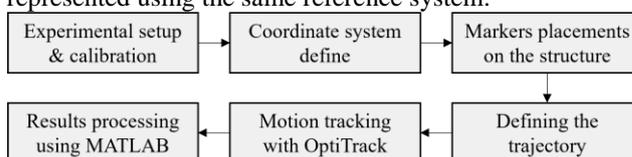

**Fig. 6.** Methodology used for accuracy of the PARA-SILSROB robot.

The repeatability of the robot was determined using the coordinate measuring system. For this experiment the evolution of the mobile platform was observed using 5 sets of measurements. The robot was moved from starting position to a selected point 5 times consecutively. Using the Stinger II measurement system, the position of the mobile platform was determined by touching three planes on the mobile platform using the measuring arm, after moving the robot to the second point, the same three planes were measured. In the end 5 sets of coordinates resulted for 4 points. After measurements, data was further processed to determine the repeatability of the robotic system.

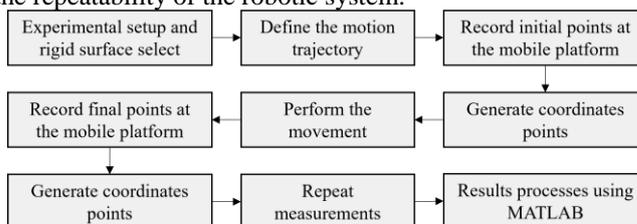

**Fig. 7.** Methodology used for repeatability of the PARA-SILSROB robot.

## 3    Results

The velocity and acceleration of the PARA-SILSROB used to determine accuracy and repeatability is 20 mm/s and 50 mm/s$^2$, these parameters are chosen according to the medical protocol and the surgeon requirements [2, 10].

The accuracy was determined for a specific trajectory used to move the mobile platform (RCM) near the insertion point and the results are presented in Fig. 8. When comparing the centroid trajectory and the trajectory obtained using the kinematic model of the robot the Root-Mean-Square Error (RMSE) on the three axes was determined. For *X* axis RMSE was 0.2866 mm, for *Y* axis RMSE was 0.4052 and for the *Z* axis the RMSE was 0.1161. Analyzing the result obtained, the accuracy of the system for this specific trajectory was determined using the RMSE between the RMSE of OptiTrack and the RMSE of values extracted from the experimental model during the experimental run and the result is 0.3 mm, this accuracy is influenced by the OptiTrack system limitation and the light sources, according to the technical requirement of this system the accuracy is 0.1 mm in ideal conditions.

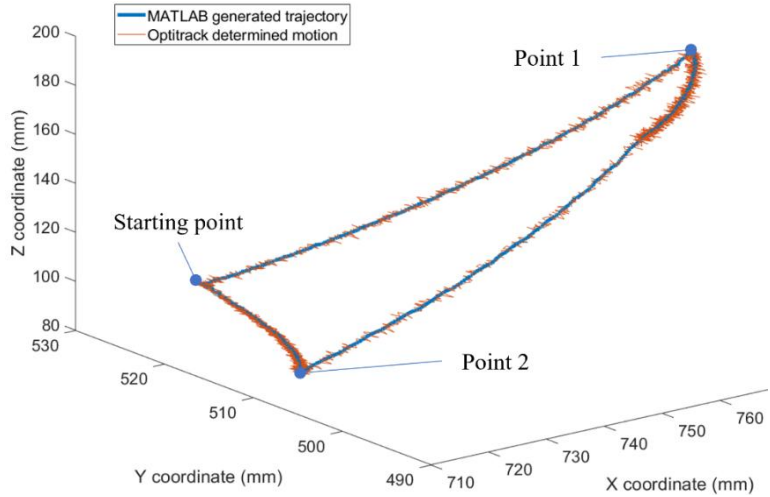

**Fig. 8.** MATLAB generated trajectory compared with measured trajectory.

$$C(x, y) = \left( \frac{x_1 + x_2 + x_3}{3}, \frac{y_1 + y_2 + y_3}{3} \right) \qquad (1)$$

The repeatability of the PARA-SILSROB was determined according to ISO 21748 [20] and based on the trajectory presented above the data was recorded using the Stinger II robotic arm and the results of data are presented in Fig. 9. In the end the repeatability for point 1 was 0.37 mm, for point 2 was 0.25mm, for point 3 was 0.31mm and for point 4 was 0.23mm. The repeatability standard deviation for the entire robotic system on the trajectory defined above is 0.18. Based on the studies presented in [21, 22] the accuracy of the robots used in surgery (da Vinci) is 1.02mm and the repeatability in classical surgery is 1.66 mm, this depends on the experience of surgeon, being also influenced by the hand tremors.

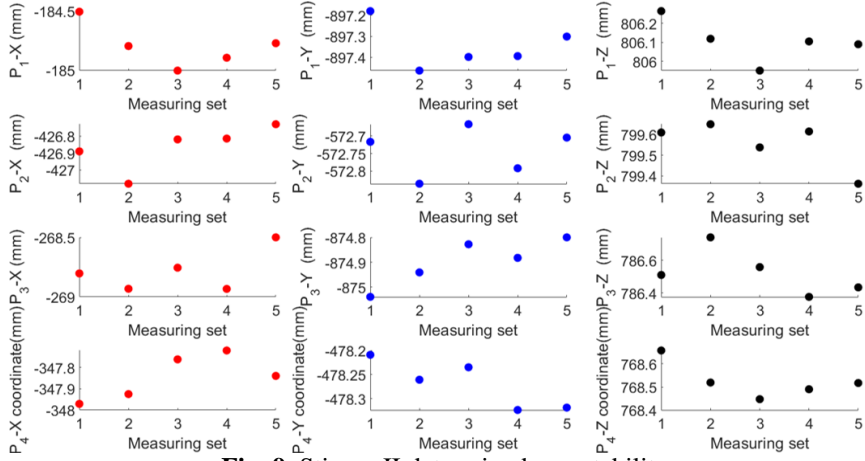

**Fig. 9.** Stinger II determined repeatability.

## 4   Conclusions

The paper presents the experimental assessment of the accuracy and repeatability for the specified trajectory (moving the mobile platform from an arbitrary position near by the insertion point) of the PARA-SILSROB robot, using OptiTrack measurement system for accuracy assessment and Stinger II measuring arm for repeatability assessment. This robot and the experimental test were developed according to the actual requirements used in this field and the medical protocol used in SILS surgery. The experimental measuring was carried out in laboratory conditions and an accuracy of 0.3 mm was obtained and a repeatability standard deviation of 0.18 was also obtained for this trajectory, the results being promising for the first iteration of the robot. Taking into account the fact that the active instruments are manipulated in the intraoperative field based on the images of the laparoscopic camera and in the case of robotic assisted surgery the margin of error is relatively higher due to the elasticity of the tissue, compared to other surgical interventions, the results obtained were considered acceptable.

Future developments will focus on extended this study for different trajectories and the experimental tests on human phantom in relevant medical conditions.

## Acknowledgements

This work was supported by the project New smart and adaptive robotics solutions for personalized minimally invasive surgery in cancer treatment - ATHENA, funded by European Union – NextGenerationEU and Romanian Government, under National Recovery and Resilience Plan for Romania, contract no. 760072/23.05.2023, code CF 116/15.11.2022, through the Romanian Ministry of Research, Innovation and Digitalization, within Component 9, investment I8 and by a grant of the Ministry of Research, Innovation and Digitization, CNCS/CCCDI—UEFISCDI, project number PN-III-P2-2.1-PED-2021-2790 694PED—Enhance, within PNCDI III.